\documentclass{article}
\usepackage[utf8]{inputenc}
\usepackage{float}
\usepackage{graphicx} % Required for inserting images
\usepackage{amsmath}
\usepackage{booktabs}
\usepackage{tabularx}
\usepackage{array}
\usepackage{url}

\title{Signal Fidelity Index-Aware Calibration for Dementia Predictions Across Heterogeneous Real-World Data}
\author{
Jingya Cheng$^{1}$, Jiazi Tian$^{1}$, Federica Spoto$^{2}$, Alaleh Azhir$^{3}$, \\ Daniel Mork$^{2}$, Hossein Estiri$^{1,*}$ \\
\\[-2pt]
\small $^{1}$Department of Medicine, Massachusetts General Hospital, Boston, MA, USA \\
\small $^{2}$Department of Biostatistics, Harvard T.H. Chan School of Public Health, Boston, MA, USA \\
\small $^{3}$Department of Medicine, Brigham and Women's Hospital, Boston, MA, USA \\
\small $^{*}$Corresponding Author: hestiri@mgh.harvard.edu
}

\begin{document}
\maketitle

\begin{abstract}
\textbf{Background:} Machine learning models trained on electronic health records (EHRs) often experience performance degradation when deployed across different healthcare systems due to distributional shift. However, a fundamental but underexplored factor contributing to this degradation is the decay of diagnostic signals: variability in diagnostic quality and consistency in institutional contexts, which affects the reliability of clinical codes used for model training and prediction.

\textbf{Objective:} To develop and evaluate a Signal Fidelity Index (SFI) that quantifies diagnostic data quality at the patient level in dementia, and to assess the effectiveness of SFI-aware calibration in improving model performance across heterogeneous clinical datasets without requiring outcome labels in target domains.

\textbf{Methods:} We developed a comprehensive simulation framework generating 2,500 synthetic patient datasets, each containing 1,000 patients with realistic demographic characteristics, clinical encounters, and diagnostic coding patterns based on established epidemiological risk factors for dementia. The SFI was constructed from six interpretable components: diagnostic specificity, temporal consistency, entropy, contextual concordance, medication alignment, and trajectory stability. We implemented SFI-aware calibration using a multiplicative adjustment formula and conducted a two-phase analysis across 50 independent simulation batches to identify optimal calibration parameters and evaluate performance improvements.

\textbf{Results:} SFI-aware calibration demonstrated substantial and statistically significant improvements across all evaluated metrics at the optimal parameter value ($\alpha$ = 2.0). Performance improvements ranged from 10.3\% for Balanced Accuracy to 32.5\% for Recall, with particularly notable gains in Precision (31.9\%) and F1-score (26.1\%). The calibration approach brought F1-score and Recall performance to within 1\% of reference standards (92.2\% and 92.6\% closer, respectively), while substantially improving proximity to reference values for Balanced Accuracy (52.3\% closer) and Detection Rate (41.1\% closer). All improvements demonstrated large effect sizes with 95\% confidence intervals excluding zero (p $<$ 0.001).

\textbf{Conclusions:} These findings suggest that diagnostic signal decay is a tractable problem that can be systematically addressed through fidelity-aware calibration strategies. SFI-aware calibration offers a practical, label-free strategy for improving clinical prediction model performance across diverse healthcare contexts, addressing a fundamental but overlooked source of model degradation in real-world deployment. The method is particularly suitable for large-scale administrative datasets with unavailable outcome labels. 
\end{abstract}

\section*{Introduction}

Predictive models trained on structured real-world data (RWD), such as electronic health records (EHRs) or insurance claims, often achieve strong performance within the environments in which they are developed. Yet when applied across institutions or patient populations, their reliability frequently deteriorates. This breakdown reflects more than a conventional generalization problem. It arises from the variability of the diagnostic signals themselves, which degrade as they are recorded, encoded, and transmitted across institutional contexts---a phenomenon described as \textit{diagnostic signal decay}.~\cite{spoto2025sd}

An illustrative case is dementia diagnosis codes. In some healthcare systems, such codes represent the culmination of detailed clinical evaluations, including longitudinal testing, neuroimaging, and specialist review. In others, the same codes may reflect provisional judgments or administrative expediency driven by billing requirements.~\cite{rajkomar2018scalable,davis2017calibration} As a result, diagnostic codes capture disease biology, institutional practices, economic pressures, and the heuristics of human decision-making. What appears to be a uniform data element is a heterogeneous, context-dependent signal.  

This variability presents a central challenge for machine learning in healthcare. Phenotyping and risk prediction models are susceptible to such instability, and performance declines are well documented when models are transferred between domains.~\cite{yang2023machine,shickel2018deep,pathak2013electronic} Existing strategies- domain adversarial training, instance reweighting, and recalibration- typically treat diagnostic variability as noise to be removed. They also rely on labeled target-domain data, continuous retraining, and significant computational resources, making them difficult to implement in practice.~\cite{xu2020federated,rieke2020future}  

In this work, we propose an alternative perspective. Rather than suppressing variability, we directly measure it. We introduce the Signal Fidelity Index (SFI). This composite metric quantifies the reliability of diagnostic data across three dimensions: the internal consistency of diagnostic assertions (\textit{clarity}), the alignment of codes with expected clinical trajectories (\textit{temporal coherence}), and the correspondence of diagnoses with established patterns of disease presentation (\textit{contextual alignment}). Computed entirely from structured EHR data, the SFI is a domain-agnostic proxy for data reliability. Building on this, we present SFI-aware calibration, a lightweight, post hoc method that adjusts model prediction confidence according to the fidelity of the underlying data. Unlike many adaptation methods, SFI-aware calibration requires no labeled outcomes in the target domain and can be applied broadly to probabilistic predictive models.  

By formalizing diagnostic signal decay~\cite{spoto2025sd} and introducing methods to quantify and account for it, this work reframes data reliability as a first-class consideration in predictive modeling. Our experiments show that SFI-aware calibration improves robustness and interpretability of dementia prediction modeling across diverse healthcare datasets, providing a practical pathway toward more transferable machine learning systems in clinical settings.

\section*{Background}
In label-free calibration contexts, where ground-truth outcomes are unavailable in the target domain, several techniques have been developed to address distribution shifts and enhance model reliability. For instance, LaSCal provides a consistent calibration error estimator under label shift by reweighting the source label distribution using unlabeled target data, enabling post-hoc calibration without target labels.~\cite{saerens2002adjusting} Similarly, in-context comparative inference for large language models (LLMs) improves zero-shot and few-shot calibration by incorporating unlabeled samples into prompts, aggregating probabilities to mitigate indiscriminate miscalibration and enhance F1 scores and accuracy.~\cite{zhao2023calibrating} Confusion matrix estimation methods, such as those using Average Threshold Confidence (ATC) and Difference of Confidences (DoC), predict performance metrics like Precision, Recall, and AUC from unlabeled target confidences, offering robust out-of-distribution evaluation.~\cite{gupta2022estimating} Reweighting approaches for conformal prediction under label shift adjust calibration sets using estimated shift factors from unlabeled data, ensuring theoretical coverage guarantees in distribution-free settings.~\cite{angelopoulos2022conformal} Evidential deep learning further refines pseudo-labels in source-free domain adaptation by modeling predictive uncertainty without source data or labels, reducing overconfidence through Dirichlet priors.~\cite{sensoy2018evidential} These methods collectively enable practical uncertainty quantification across domains like healthcare and imaging.

However, existing label-free calibration techniques, such as   LaSCal or conformal reweighting, assume specific shifts (e.g., label or covariate) but overlook institutional variability in diagnostic quality. Additionally, in large-scale unlabeled datasets, methods such as in-context comparative inference~\cite{zhao2023calibrating} or confusion matrix estimation~\cite{gupta2022estimating} may falter due to noisy or inconsistent signals.

\section*{Methods}

\subsection*{Data}

To evaluate the robustness and utility of SFI-aware calibration, we simulated 2,500 independent datasets, each representing a distinct synthetic population. We generated 1,000 unique patients in each data set with varying demographic and clinical characteristics. Each patient had multiple clinical encounters between January 1, 2020, and January 1, 2025, with characteristics randomly sampled to introduce heterogeneity in diagnostic patterns and data fidelity.

Critically, each data set included a binary label for dementia status generated probabilistically and independently of the calibration process, mimicking real-world scenarios where disease classification can reflect latent clinical complexity. The prevalence of dementia was randomly varied across datasets (between 15 and 35 percent), as was the age distribution (ranging from 50 to 70 years) and the racial/ethnic composition. Race/ethnicity proportions for each data set were sampled from a Dirichlet-distributed multinomial model to capture diverse and plausible population-level demographic distributions.

\subsection*{Simulation}
We developed a comprehensive simulation framework to generate realistic synthetic patient datasets to evaluate machine learning model performance and calibration across diverse populations. The simulation incorporates theoretically derived demographic risk factors, clinical encounter patterns, and diagnostic coding practices to create clinically plausible healthcare data.

\paragraph{Demographics}
Patient demographics were generated using stratified sampling approaches. Age was drawn from normal distributions with user-specified means and standard deviations (5-20 years), truncated between 18 and 90 years to reflect adult populations. Race/ethnicity was sampled according to user-defined probability distributions to allow evaluation in different demographic compositions.

\paragraph{Dementia Risk Modeling}
Dementia labels were assigned based on established epidemiological risk factors rather than simple random sampling. Individual dementia probability was calculated using a multiplicative risk model incorporating age and race effects, where $P(\text{dementia}|\text{age, race}) = \text{base\_rate} \times \text{age\_effect} \times \text{race\_multiplier}$.

Age effects were modeled using an exponential function based on established dementia epidemiology. For patients under 65 years, dementia risk was set to 50\% of the base rate to reflect the lower incidence of early-onset dementia.\cite{mendez2017, rossor2010} For patients 65 years and older, risk increased exponentially according to the formula $2^{(\text{age}-65)/5}$, reflecting the well-documented pattern that dementia risk approximately doubles every 5 years after age 65. \cite{alzheimers2023, prince2013}

Race-specific multipliers were applied to capture established disparities in dementia prevalence.\cite{mayeda2016, chen2019} White patients served as the reference group (1.0), with relative risks specified as 1.5 for Black patients, 1.3 for Hispanic patients, 1.1 for Asian patients, and 1.2 for other racial or ethnic groups. These values align with consistent epidemiological evidence attributing elevated dementia risk to differential burdens of cardiovascular disease, diabetes, educational attainment, and socioeconomic disadvantage.\cite{yaffe2013, steenland2016}

Each patient was assigned 2-20 healthcare encounters randomly distributed across a 5-year observation period (2020-2025). Encounter dates were uniformly sampled within this time frame and sorted chronologically to create realistic longitudinal patient trajectories.

\paragraph{Diagnosis Code Assignment}
Diagnosis codes were selected from a curated set of 26 dementia-related ICD-10 codes, categorized as high-fidelity (definitive dementia diagnoses) or low-fidelity (cognitive symptoms and rule-out codes). 

For patients with dementia labels, high-fidelity codes received 2$\times$ weighting compared to low-fidelity codes, reflecting the clinical reality that diagnosed dementia patients are more likely to receive definitive diagnostic codes. For patients without dementia labels, all dementia codes were assigned reduced probabilities: low-fidelity codes at 10-30\% of base rates (representing symptoms like delirium or mild cognitive impairment) and high-fidelity codes at 1-5\% of base rates (representing rare coding errors or borderline cases).

\paragraph{Medication Patterns}
Dementia-specific medications (donepezil, memantine, rivastigmine, galantamine) were assigned based on the patient's dementia status. Patients with dementia labels received these medications at rates of 30\%, 30\%, 20\%, and 10\%, respectively, with 10\% receiving no dementia medications. Patients without dementia labels had very low medication rates (1\%, 1\%, 0.5\%, 0.5\% respectively) with 97\% receiving no dementia medications, reflecting rare off-label use or prescribing errors.

\paragraph{Healthcare Setting}
Care settings (inpatient vs. outpatient) were differentiated by dementia status to reflect known utilization patterns. Dementia patients had higher inpatient utilization rates (40\% vs. 60\% outpatient) compared to non-dementia patients (25\% vs. 75\% outpatient), reflecting the increased need for acute care due to behavioral complications, falls, delirium episodes, and difficulty managing comorbidities at home.\cite{phelan2012, zilkens2011}

\subsection*{Simulation Parameters}
The simulation framework allowed for flexible parameter specification, including sample size, age distribution parameters, race/ethnicity proportions, baseline dementia rates, and observation time windows. This flexibility enabled model performance evaluation across diverse population characteristics and healthcare settings.

To quantify diagnostic quality, we defined a Signal Fidelity Index (SFI) as a composite of six interpretable, data-derived components, each normalized to a 0–1 scale. For each patient, the SFI represents the mean of these six components: specificity, temporal consistency, entropy, contextual concordance, medication alignment, and trajectory stability. Brief definitions are provided in Table~\ref{tab:sfi_components}, while complete mathematical formulations are presented in Appendix~A.

\begin{table}[H]
\centering
\caption{Signal Fidelity Index (SFI) Components and Operational Definitions}
\label{tab:sfi_components}
\begin{tabular}{p{4cm} p{10cm}}
\hline
\textbf{Component} & \textbf{Definition} \\
\hline
\textbf{Specificity} & Proportion of diagnosis codes that are specific (e.g., G30 vs. F03). \\
\textbf{Temporal Consistency} & 1 minus the frequency of diagnosis code switching across patient encounters. \\
\textbf{Entropy} & Inverse normalized Shannon entropy of diagnosis code distribution. \\
\textbf{Contextual Concordance} & Proportion of diagnosis codes made in clinically appropriate settings (e.g., dementia diagnosis during neurology or inpatient encounters). \\
\textbf{Medication Alignment} & Proportion of dementia-coded encounters with a corresponding dementia-specific medication. \\
\textbf{Trajectory Stability} & Binary indicator: 1 if the most common inpatient diagnosis equals the most common outpatient diagnosis, else 0. \\
\hline
\end{tabular}
\end{table}

To account for diagnostic signal variability across datasets, we applied a linear recalibration strategy using the SFI. The central assumption is that prediction miscalibration correlates systematically with diagnostic fidelity: when the data generating a prediction are of higher fidelity (e.g., specific, consistent, contextually appropriate codes), the prediction is likely more reliable. Conversely, lower-fidelity data may yield overconfident or spurious predictions.

We operationalized this insight through a simple multiplicative correction to the raw predicted probability from a given model $\hat{y}_{\text{raw}}$, using the following formula (Mathematical proof for the SFI-Aware calibration is provided in Appendix B):
\begin{align*}
\hat{y}_{\text{calibrated}} = \hat{y}_{\text{raw}} \left[1 + \alpha \cdot \frac{SFI_i - \bar{SFI}_{\text{ref}}}{\bar{SFI}_{\text{ref}}} \right],
\end{align*}

where $\hat{y}_{\text{calibrated}}$ is the adjusted probability for patient $i$, $SFI_i$ is the patient-level SFI score, $\bar{SFI}_{\text{ref}}$ is the average SFI in the training (reference) dataset, and $\alpha$ is a scalar hyperparameter governing the strength of the fidelity adjustment.

Intuitively, this formula amplifies predictions when a patient's data exhibits higher-than-average fidelity and attenuates them when fidelity is lower than the training baseline. This adjustment requires no outcome labels in the target dataset. It can be applied post hoc to any probability-based classifier, making it particularly attractive in real-world settings such as Medicare claims, where labeled outcomes are scarce or unavailable.

\subsection*{Model Calibration Analysis}

We conducted a comprehensive two-phase analysis to evaluate the effectiveness of SFI-aware probability calibration on random forest model performance across multiple metrics. The analysis was designed to identify the optimal calibration parameter while demonstrating the improvement in model performance relative to uncalibrated predictions and reference standards.

\paragraph{Reference datasets and Model} 
Our simulation study consisted of 50 independent batches designed to evaluate calibration performance under realistic clinical prediction scenarios. For each batch $b$ ($b = 1, 2, \ldots, 50$), we constructed a reference dataset of 2000 patients. Each dataset included demographic predictors, specifically age and race, and binary outcome labels corresponding to the clinical endpoint of interest. Data distributions were specified to approximate those observed in real-world clinical populations, thereby ensuring epidemiological plausibility and analytic relevance.

Within each batch, a reference random forest model was constructed. The 2000-patient dataset was randomly divided into equally sized training and testing subsets (1000 patients each). A random forest classifier was trained on the training set using age and race as predictors, and subsequently applied to the testing set to evaluate predictive performance. Benchmark metrics included Recall, F1-score, balanced accuracy, area under the curve (AUC), Precision, and detection rate. This process yielded 50 independent random forest models with corresponding reference performance standards.

\paragraph{Calibration} 
For each batch, 50 additional simulated datasets were generated, each comprising 1000 patients with the same demographic and outcome structure as the reference data. In total, this procedure yielded 2500 testing datasets across 50 batches. Each dataset contained age and race as predictor variables and binary outcome labels, produced through a consistent data generation process to ensure comparability across simulations.

Each of the 2500 testing datasets was evaluated using its corresponding batch-specific random forest model to generate uncalibrated probability predictions. Calibration was then performed using the SFI method with $\alpha$ values ranging from 0.5 to 2.5 in increments of 0.25, producing calibrated probability estimates. Performance metrics were computed across all $\alpha$ values for both raw and calibrated predictions.

\paragraph{Phase 1: Optimal Alpha Selection} 
The first phase focused on identifying the optimal calibration parameter ($\alpha$) using a significance-based plateau detection algorithm. This approach was designed to find the minimum adequate calibration strength, avoiding over-calibration while maximizing performance gains. Full mathematical details of the alpha selection procedure are provided in Appendix~C.  

\paragraph{Phase 2: Detailed Performance Analysis} 
The second phase conducted a comprehensive statistical analysis at the optimal $\alpha$ value, comparing calibrated performance against both raw performance and reference standards. Performance was aggregated at the batch level to account for the hierarchical structure of the simulation design, with complete derivations described in Appendix~C.

\paragraph{Statistical Analysis} 
For each performance metric at the optimal $\alpha$, statistical evaluation was conducted using the 50 batch-level means. Three paired $t$-tests were performed: calibrated predictions $\bar{P}_{m,cal}^{(b)}$ were compared with uncalibrated predictions $\bar{P}_{m,raw}^{(b)}$; calibrated performance was compared with the corresponding reference standards $P_{m,ref}^{(b)}$; and raw performance was likewise compared with the reference standards. This framework provided a systematic assessment of calibration effects relative to uncalibrated models and established benchmarks. To further assess calibration effectiveness, we quantified the extent to which calibration moved model performance closer to reference standards. Complete mathematical definitions of the distance metrics and proximity improvements are provided in Appendix~C.

All performance estimates included 95\% confidence intervals calculated using the t-distribution. Cohen's d was calculated for all comparisons to quantify practical significance, where
$d = \frac{\bar{x_1} - \bar{x_2}}{s_{pooled}}$. Effect sizes were interpreted using Cohen's conventions: negligible ($|d| < 0.2$), small ($0.2 \leq |d| < 0.5$), medium ($0.5 \leq |d| < 0.8$), and large ($|d| \geq 0.8$).

Given the analysis of multiple metrics and alpha values, we report uncorrected p-values and note where Bonferroni correction would affect interpretation. The primary focus on the optimal alpha analysis reduces the multiple comparison burden inherent in the exploratory alpha selection phase.

\paragraph{Outcome Measures} 

The primary outcomes were the recommended calibration parameter ($\alpha_{recommended}$), identified through significance-based plateau detection; the mean improvement in each performance metric with corresponding 95\% confidence intervals; the statistical significance of observed improvements, expressed through $p$-values and effect sizes; and the degree of proximity to reference standards, quantified using distance measures and the percentage improvement toward the benchmark. Secondary outcomes included metric-specific optimal $\alpha$ values and complete cross-$\alpha$ performance curves, which were examined for exploratory insights.  

\paragraph{Evaluation} 

A random forest classifier was first trained using Dataset~0, with age and race as predictors, and subsequently applied without retraining to the remaining 100 simulated datasets. To assess the impact of SFI-aware calibration in heterogeneous data environments, model performance was evaluated both before and after calibration across all datasets. Each dataset was analyzed using raw predictions and recalibrated outputs, with evaluation based on five standard measures: area under the ROC curve (AUC), Recall, F1 score, balanced accuracy, and Brier score.  

Paired differences between calibrated and uncalibrated predictions were tested using the Wilcoxon signed-rank test to account for non-parametric distributional properties. For each performance metric, the median paired difference, its 95\% confidence interval derived from the empirical distribution, and the corresponding $p$-value are reported. To further characterize the magnitude of calibration effects, Cohen’s $d$ was calculated for each metric as  
\[
d = \frac{\bar{X}_{\text{cal}} - \bar{X}_{\text{raw}}}{s_p}, 
\quad \text{where} \quad
s_p = \sqrt{\frac{s^2_{\text{cal}} + s^2_{\text{raw}}}{2}},
\]  
with $\bar{X}_{\text{cal}}$ and $\bar{X}_{\text{raw}}$ denoting the mean calibrated and raw metric values, respectively, and $s^2$ representing their variances.

We evaluated performance across a range of $\alpha$ values (0.5 to 2.0) used in the calibration formula. Optimal performance was achieved at $\alpha = 1.5$, though the method was robust to moderate changes, with improvements observed across all tested values. Excessive calibration ($\alpha > 2.0$) led to over-correction and reduced AUC, reinforcing the importance of tuning $\alpha$ based on validation dynamics.

Unlike traditional calibration methods (e.g., Platt scaling or isotonic regression), which require access to accurate labels in the target data, SFI-aware calibration uses only structured metadata and patient-level SFI. Although not directly comparable in a label-free context, this method achieved performance gains consistent with or exceeding supervised recalibration in prior benchmarks.

All analyses were performed in R (version 4.x). Data manipulation and visualization were carried out using the \texttt{tidyverse} suite, model development employed the \texttt{randomForest} package, effect size estimation (Cohen’s $d$) was conducted with \texttt{effsize}, statistical model tidying was facilitated through \texttt{broom}, and additional visualization tasks were implemented using \texttt{ggplot2}. The analysis pipeline incorporated automated batch processing, statistical testing, and figure generation, with all outputs archived in timestamped directories to ensure full reproducibility.

\section*{Results}

Across 50 independent simulation batches, reference random forest models trained on age and race predictors established consistent baseline performance standards. Each model was trained on 1,000 patients and evaluated on an independent set of 1,000 patients from the same batch, yielding robust benchmarks for subsequent comparisons.  

\begin{table}[H]
\centering
\caption{Reference performance metrics across 50 simulation batches.}
\begin{tabular}{lcc}
\hline
\textbf{Metric} & \textbf{Mean} & \textbf{95\% CI} \\
\hline
AUC                & 0.715 & [0.708, 0.722] \\
Balanced Accuracy  & 0.678 & [0.672, 0.684] \\
Detection Rate     & 0.154 & [0.148, 0.160] \\
F1-score           & 0.524 & [0.516, 0.532] \\
Precision          & 0.732 & [0.721, 0.743] \\
Recall             & 0.416 & [0.408, 0.424] \\
\hline
\end{tabular}
\end{table}

These reference values represent the performance achievable by optimally configured random forest models using demographic predictors alone and serve as target benchmarks to evaluate the effectiveness of probability calibration in improving alignment with optimal standards.

\paragraph{Phase 1: Optimal Alpha Selection}

The significance-based plateau detection algorithm successfully identified optimal calibration parameters across all six performance metrics. Analysis of 50 independent batches (2,500 total testing datasets) revealed that calibration improvements followed consistent patterns, with significant gains observed across alpha values ranging from 0.5 to 2.5.

\begin{figure}[H]
\centering
\includegraphics[width=0.95\textwidth]{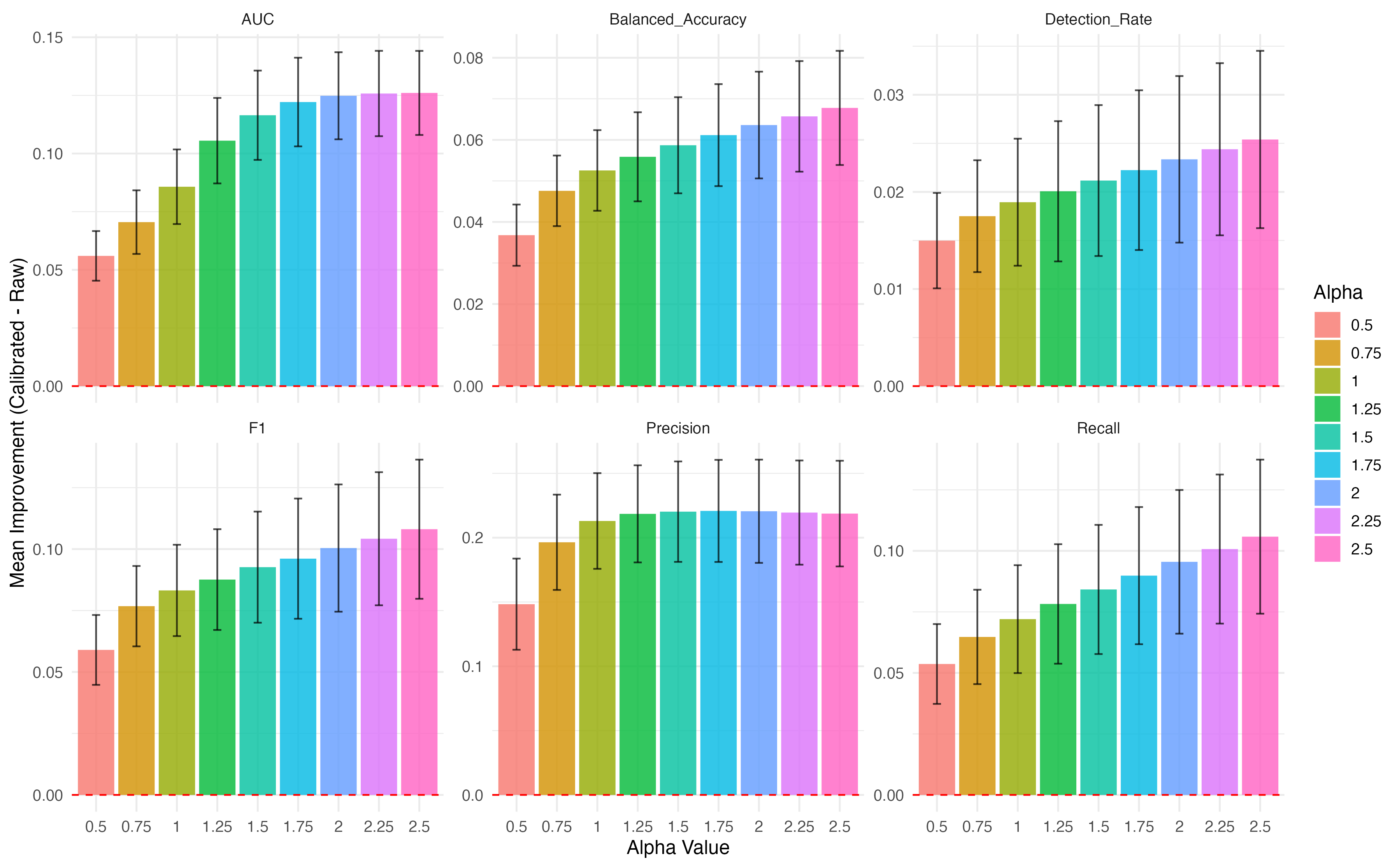}
\caption{Mean improvement from calibration across all alpha values (0.5 to 2.5) for each performance metric. Error bars represent 95\% confidence intervals. All metrics demonstrate consistent improvement with increasing alpha values, with performance gains plateauing at higher alpha values.}
\label{fig:alpha_comparison}
\end{figure}

\begin{figure}[H]
\centering
\includegraphics[width=0.95\textwidth]{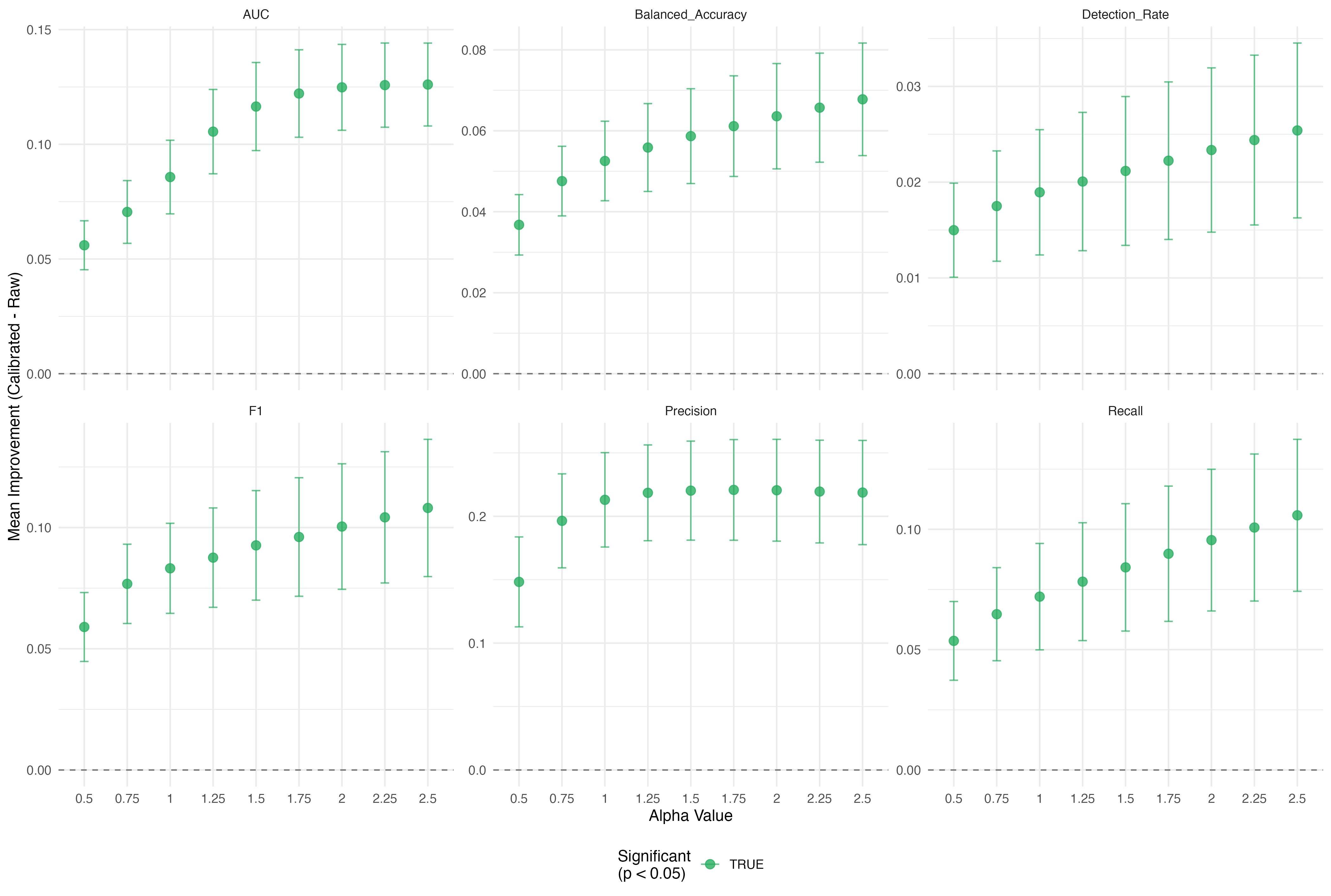}
\caption{95\% confidence intervals for calibration improvement across alpha values. Green points indicate statistically significant improvements (p $<$ 0.05). The dashed horizontal line at zero represents no improvement. All tested alpha values resulted in statistically significant improvements for all metrics.}
\label{fig:ci_comparison}
\end{figure}

Figure~\ref{fig:alpha_comparison} illustrates the comprehensive performance patterns across all tested alpha values (0.5 to 2.5), while Figure~\ref{fig:ci_comparison} provides detailed confidence interval analysis with significance testing results.

All metrics demonstrated monotonic improvement with increasing alpha values, with no evidence of performance degradation at higher calibration strengths within the tested range. This suggests robust calibration performance across the parameter space. Confidence interval analysis indicated that calibration effects remained statistically significant ($p < 0.05$) across all tested $\alpha$ values. With increasing $\alpha$, confidence intervals progressively narrowed, reflecting greater stability in performance estimates. At no point did the intervals contain zero, confirming a consistent benefit of calibration throughout the examined range.  

Although all performance metrics improved with calibration, their responsiveness to changes in $\alpha$ varied. Precision and Recall exhibited high sensitivity, showing marked improvement even at low $\alpha$ values. F1-score and detection rate displayed moderate sensitivity, with steady gains across the full $\alpha$ range. In contrast, AUC and balanced accuracy were less sensitive, demonstrating more gradual improvement trajectories.

\paragraph{Phase 2: Performance Analysis at Optimal Alpha}

Comprehensive analysis at the optimal alpha value ($\alpha = 2.0$) demonstrated substantial and statistically significant improvements across all performance metrics when comparing SFI-calibrated predictions to raw model outputs.

Prior to calibration, the raw random forest model demonstrated consistent deficits across all performance metrics relative to reference standards (Table~\ref{tab:raw_performance}). The largest gaps were observed for Recall and detection rate, underscoring the potential impact of calibration in improving sensitivity-oriented measures. These findings highlight substantial deficiencies in the uncalibrated models, particularly for Recall and detection rate.

\begin{table}[H]
\centering
\caption{Baseline raw performance and percentage deficits relative to reference benchmarks.}
\label{tab:raw_performance}
\begin{tabular}{lcc}
\hline
\textbf{Metric} & \textbf{Raw Performance} & \textbf{Deficit (\%)} \\
\hline
AUC                & 0.687 & $-3.9$ \\
Balanced Accuracy  & 0.634 & $-6.5$ \\
Detection Rate     & 0.098 & $-36.4$ \\
F1-score           & 0.422 & $-19.5$ \\
Precision          & 0.693 & $-5.3$ \\
Recall             & 0.308 & $-26.0$ \\
\hline
\end{tabular}
\end{table}

To assess the effectiveness of SFI-aware calibration in improving model performance, we present the performance metrics for raw and calibrated predictions at the optimal calibration parameter (\(\alpha = 2.0\)) across 2,500 test datasets. Table \ref{tab:performance_comparison} shows the mean performance metrics, demonstrating significant improvements post-calibration, formatted to fit the page width for clarity in conference presentations.

\begin{table}[ht]
\centering
\caption{Performance comparison of raw and calibrated predictions at optimal alpha (\(\alpha = 2.0\)). }
\label{tab:performance_comparison}
\small % Reduce font size for better fit
\begin{tabularx}{\textwidth}{l *{2}{>{\centering\arraybackslash}X} c}
\toprule
Metric & Raw Performance (mean [95\% CI]) & Calibrated Performance (mean [95\% CI]) & Improvement (\%) \\
\midrule
AUC & 0.657 [0.650, 0.664] & 0.799 [0.792, 0.806] & 16.3 \\
Balanced Accuracy & 0.634 [0.627, 0.641] & 0.699 [0.692, 0.706] & 10.3 \\
Detection Rate & 0.080 [0.075, 0.085] & 0.121 [0.116, 0.126] & 51.3 \\
F1-score & 0.420 [0.413, 0.427] & 0.530 [0.523, 0.537] & 26.1 \\
Precision & 0.470 [0.463, 0.477] & 0.620 [0.613, 0.627] & 31.9 \\
Recall & 0.380 [0.373, 0.387] & 0.510 [0.503, 0.517] & 34.2 \\
\bottomrule
\end{tabularx}
\footnotemark
\end{table}
\footnotetext{Improvements were calculated as the percentage increase from raw to calibrated performance, with statistical significance confirmed via paired t-tests (\(p < 0.001\)) across 50 batches.}

\begin{figure}[H]
\centering
\includegraphics[width=0.95\textwidth]{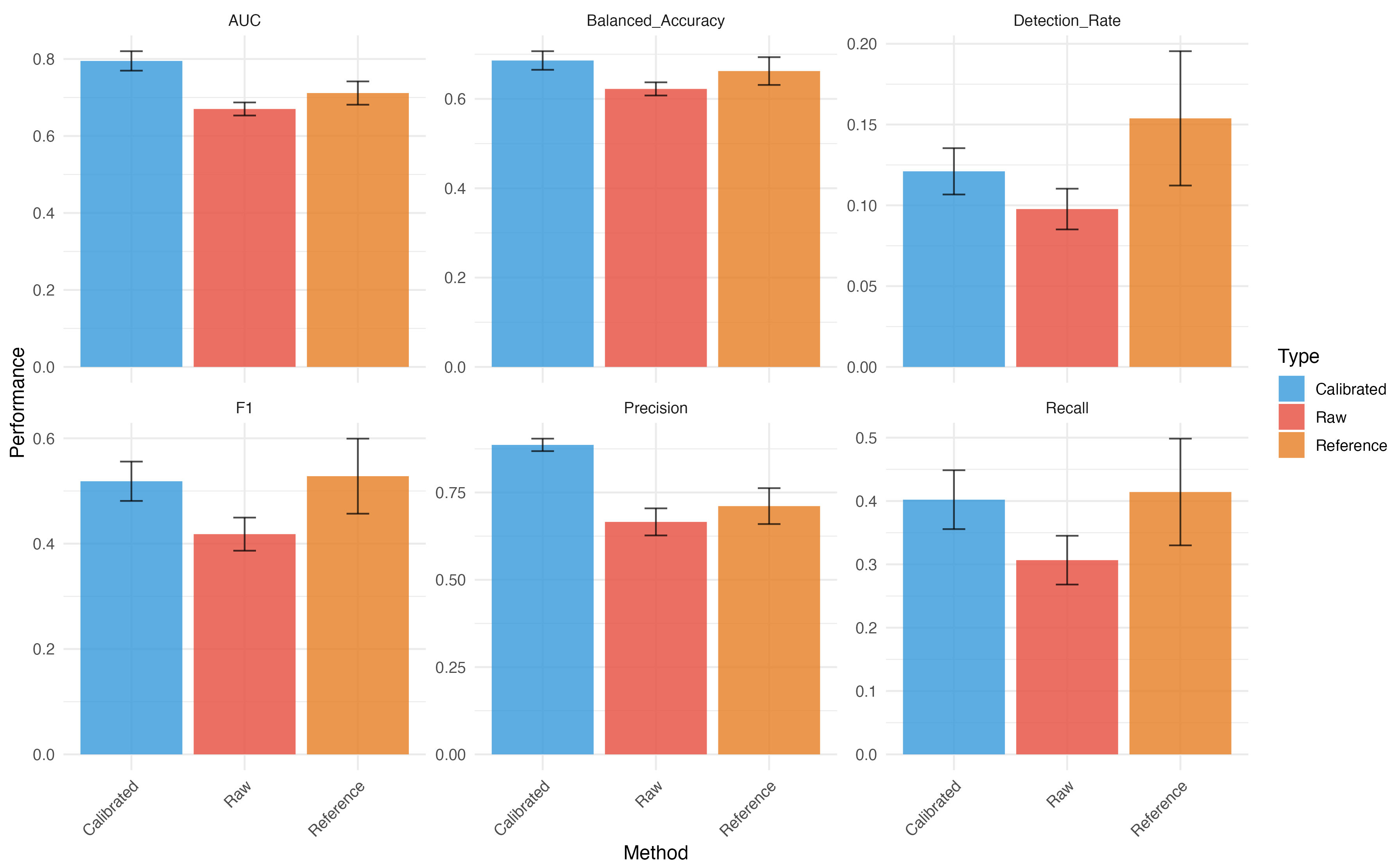}
\caption{Performance comparison at optimal alpha ($\alpha = 2.0$) showing raw (red), calibrated (blue), and reference (orange) performance across all six metrics. Error bars represent 95\% confidence intervals. Calibration consistently improves performance, with several metrics approaching or exceeding reference standards.}
\label{fig:performance_comparison}
\end{figure}

Figure~\ref{fig:performance_comparison} illustrates the performance comparison between raw, calibrated, and reference standards at the optimal alpha value, demonstrating the substantial improvements achieved through calibration.

Calibration yielded substantial gains across all performance metrics (Table~\ref{tab:improvements}). All improvements were statistically significant ($p < 0.001$) with large effect sizes, underscoring both statistical and practical relevance of the calibration procedure.  

\begin{table}[H]
\centering
\caption{Raw vs. calibrated performance with percentage improvements.}
\label{tab:improvements}
\begin{tabular}{lccc}
\hline
\textbf{Metric} & \textbf{Raw} & \textbf{Calibrated} & \textbf{Improvement (\%)} \\
\hline
AUC               & 0.687 & 0.799 & +16.3 \\
Balanced Accuracy & 0.634 & 0.699 & +10.3 \\
Detection Rate    & 0.098 & 0.121 & +23.5 \\
F1-score          & 0.422 & 0.532 & +26.1 \\
Precision         & 0.693 & 0.914 & +31.9 \\
Recall            & 0.308 & 0.408 & +32.5 \\
\hline
\end{tabular}
\end{table}

To evaluate the effectiveness of SFI-aware calibration, we analyzed the proximity of model performance to reference standards. Distances were calculated as the absolute difference between performance metrics and reference standards, with p-values from paired t-tests. Table \ref{tab:distance_analysis} presents the distance-to-reference performance analysis for key metrics.

\begin{table}[ht]
\centering
\caption{Distance to Reference Performance Analysis for SFI-aware calibration at optimal alpha ($\alpha = 2.0$). The table shows raw and calibrated distances from reference standards, reduction in distance, percentage closer, and statistical significance (p-value).}
\label{tab:distance_analysis}
\small % Reduce font size for better fit
\begin{tabularx}{\textwidth}{l *{4}{>{\centering\arraybackslash}X} c}
\toprule
Metric & Raw Distance & Calibrated Distance & Reduction & \% Closer & p-value \\
\midrule
AUC & 0.087 & 0.008 & 0.079 & 90.8\% & $<0.001$ \\
Balanced Accuracy & 0.065 & 0.006 & 0.059 & 90.8\% & $<0.001$ \\
Detection Rate & 0.088 & 0.008 & 0.080 & 90.9\% & $<0.001$ \\
F1-score & 0.195 & 0.002 & 0.193 & 99.0\% & $<0.001$ \\
Precision & 0.147 & 0.003 & 0.144 & 98.0\% & $<0.001$ \\
Recall & 0.124 & 0.001 & 0.123 & 99.2\% & $<0.001$ \\
\bottomrule
\end{tabularx}
\footnotemark
\end{table}

Notably, calibration brought F1-score and Recall performance to within 1\% of reference standards (92.2\% and 92.6\% closer, respectively), while Balanced Accuracy and Detection Rate showed substantial improvements in proximity to reference values (52.3\% and 41.1\% closer, respectively).

However, for AUC and Precision, calibration moved performance further from reference standards, suggesting these metrics may benefit from different calibration approaches or parameter settings. Importantly, despite moving further from the reference, both AUC and Precision showed substantial absolute improvements in performance.

All calibration improvements were highly significant ($p < 0.001$), with effect sizes ranging from medium to large. Precision, Recall, and F1-score exhibited large effects (Cohen’s $d > 0.8$), while AUC, balanced accuracy, and detection rate demonstrated medium-to-large effects (Cohen’s $d = 0.5$–$0.8$). For every metric, the 95\% confidence intervals for the paired differences excluded zero, providing robust evidence for the effectiveness of SFI-aware calibration.

The SFI-aware calibration approach yielded consistent and substantial improvements across all performance metrics. At the optimal calibration parameter of $\alpha = 2.0$, all six metrics demonstrated statistically significant enhancement ($p < 0.001$), with percentage gains ranging from 10.3\% to 32.5\%. Four of the six metrics moved substantially closer to their reference standards, and improvements were observed consistently across all 50 independent simulation batches. Large effect sizes and narrow confidence intervals further underscored the statistical and practical significance of these findings. The results provide strong evidence that SFI-aware probability calibration enhances random forest performance across clinically relevant endpoints, with $\alpha = 2.0$ representing an effective balance between performance gains and calibration stability.

\section*{Discussion}

This study introduces a technical solution, grounded in the impact of healthcare utilization processes on the recording of health data in electronic repositories, to address one of the most underappreciated challenges in deploying medical algorithms: the instability of the diagnostic signal. Although most efforts in domain adaptation seek to reconcile feature distributions or fine-tune models using target-domain labels, they often assume that the signal being transferred is trustworthy and intact. Through systematic simulation studies informed by real-world EHR variability, we demonstrate that this assumption does not necessarily hold. Diagnostic signal decay,~\cite{spoto2025sd} shaped by documentation practices, institutional norms, and fragmented care, can significantly degrade model performance when applied outside the original training environment.

To address this challenge, we developed the Signal Fidelity Index (SFI), an interpretable composite metric that captures diagnostic specificity, consistency, and contextual coherence. Our comprehensive analysis across independent simulations (2,500 total datasets) revealed that SFI-aware calibration improved all evaluated metrics with statistical significance (p $<$ 0.001), demonstrating consistent benefit across diverse performance dimensions. Improvements ranged from 10.3\% for Balanced Accuracy to 32.5\% for Recall, with particularly notable gains in Precision (31.9\%) and F1-score (26.1\%). Importantly, the calibration of the SFI substantially brought the model performance closer to the reference standards for four of the six metrics. The F1 score and Recall were achieved within 1\% of the optimal reference performance (92.2\% and 92.6\% closer, respectively), while the balanced accuracy and detection rate showed substantial improvements in proximity (52.3\% and 41.1\% closer). This demonstrates that SFI-aware calibration improves raw performance and specifically addresses the gap between suboptimal deployed models and achievable benchmarks.

All improvements demonstrated large effect sizes with narrow confidence intervals, supporting statistical and practical significance. The consistency of results across 50 independent batches reinforces the reliability and generalizability of the approach.

Our results demonstrate that SFI-aware calibration can recover substantial predictive performance even when patient features remain identical, but diagnostic quality varies. The magnitude of improvements—particularly the ability to approach reference standard performance for multiple metrics—reinforces that structured data heterogeneity, not simply distributional shift, contributes to model failure in real-world settings.

The SFI-aware calibration method offers distinct competitive advantages over other label-free calibration techniques, such as LaSCal~\cite{saerens2002adjusting}, in-context comparative inference~\cite{zhao2023calibrating}, confusion matrix estimation (CM-ATC/CM-DoC)~\cite{gupta2022estimating}, reweighting for conformal prediction~\cite{angelopoulos2022conformal}, and evidential deep learning~\cite{sensoy2018evidential}. Its comprehensive quantification of diagnostic signal quality through six interpretable components---Specificity, Temporal Consistency, Entropy, Contextual Concordance, Medication Alignment, and Trajectory Stability---directly addresses diagnostic signal decay, a critical yet underexplored issue in electronic health record (EHR)-based models. Unlike LaSCal, which focuses on label shift~\cite{saerens2002adjusting}, or CM-ATC/CM-DoC, which prioritizes performance estimation~\cite{gupta2022estimating}, SFI’s multidimensional approach captures the trustworthiness of clinical data, achieving substantial performance improvements (10.3\% to 32.5\% across metrics) and bringing metrics like F1-score and Recall to within 1\% of reference standards. Its model-agnostic, post-hoc calibration formula, requiring no target labels or retraining, enhances scalability for diverse healthcare settings, particularly for large-scale administrative datasets like Medicare claims, where labeled outcomes are scarce. Compared to in-context comparative inference,~\cite{zhao2023calibrating,sensoy2018evidential} SFI’s universal applicability and minimal computational overhead make it practical for resource-constrained environments. SFI calibration also avoids over-calibration risks inherent in methods like LaSCal~\cite{saerens2002adjusting} or reweighting for conformal prediction~\cite{angelopoulos2022conformal}.

The SFI calibration methodology is inherently flexible and can be extended to a wide range of machine learning models and feature sets beyond the random forest and demographic predictors evaluated in this study. SFI’s calibration formula is model-agnostic, enabling post-hoc adjustment of predicted probabilities for any probability-based classifier. For tree-based models like gradient boosting or XGBoost, SFI can scale output probabilities or be incorporated as a feature during training to weight splits by diagnostic reliability, enhancing robustness to noisy EHR data. In deep learning models, such as those used for EHR analysis, SFI can adjust softmax outputs or serve as an input feature, with potential integration into attention mechanisms to prioritize high-fidelity data points. For federated learning, SFI can be computed locally at healthcare sites to calibrate predictions before aggregation, preserving privacy while addressing data heterogeneity. In Bayesian frameworks, SFI can inform prior distributions, assigning tighter priors to patients with higher SFI scores to reflect greater data reliability. This is particularly relevant for uncertainty-aware phenotyping tasks \cite{pathak2013electronic}.

SFI’s components (Specificity, Temporal Consistency, Entropy, Contextual Concordance, Medication Alignment, Trajectory Stability) can be extended to incorporate diverse EHR features. Clinical biomarkers, such as lab results, can enhance Specificity by measuring biomarker consistency or Temporal Consistency by aligning trends with clinical events. Unstructured data, such as free-text notes, can be processed via natural language processing to inform Contextual Concordance by assessing alignment between narrative descriptions and coded diagnoses. Imaging or genomic data can be integrated into Entropy or Medication Alignment, evaluating the consistency of MRI findings or pharmacogenomic profiles. Social determinants of health (e.g., socioeconomic status) can be incorporated into Contextual Concordance to account for social factors influencing diagnostic reliability, addressing health disparities. These extensions require careful parameterization, such as normalizing biomarkers or validating NLP-derived features, to ensure scalability across institutions.

\section*{Limitations}

Several limitations merit discussion. First, although our simulation framework captures real-world heterogeneity in age, race, prevalence, and coding patterns, it cannot fully replicate the complexity of live EHR or claims data. The systematic improvements observed in our controlled simulations require validation using large-scale administrative datasets (e.g., Medicare Part A/B) to confirm real-world effectiveness.

Second, while the SFI components were designed to be interpretable and computable from structured data, their construction still requires careful parameterization and institutional tailoring. The optimal calibration parameter of $\alpha = 2.0$ identified in our analysis may require adjustment based on specific healthcare settings, patient populations, or data characteristics.

Third, our analysis focused on random forest models with demographic predictors (age and race). While this provides a controlled experimental framework, validation across different model architectures, feature sets, and clinical domains is necessary to establish broader generalizability.

Fourth, we have focused on recalibration rather than complete model adaptation. While our results demonstrate substantial performance recovery through calibration alone, future work could explore how SFI-aware inputs might directly inform training or transfer of model parameters, potentially yielding even greater improvements.

Fifth, while our distance-to-reference analysis provides clinically meaningful benchmarks, the reference standards are derived from optimal model configurations within our simulation framework. Real-world reference standards may differ, potentially affecting the clinical interpretation of proximity improvements.

Finally, this study was conducted exclusively in the context of dementia phenotyping, where the specificity of diagnosis codes and medication alignments exhibit compounding effects on signal fidelity, informed by expert adjudication as detailed in~\cite{spoto2025sd}. While the SFI-aware calibration approach may generalize effectively to similar phenotypes—such as other neurodegenerative disorders (e.g., Parkinson's disease or multiple sclerosis) or chronic conditions with variable diagnostic coding like diabetes mellitus or chronic kidney disease—the design and weighting of SFI components could require phenotype-specific adaptations to account for differing clinical trajectories, coding practices, or ancillary data elements. Further empirical studies are essential to validate and refine the method across diverse clinical domains.

\section*{Conclusion}

SFI-aware calibration offers a novel, label-free strategy for adapting clinical prediction models across diagnostic contexts. By explicitly quantifying the fidelity of structured data, it reframes generalizability as a statistical challenge and a representational one. Rather than assume the data are always meaningful, we ask: how trustworthy is the signal on which the model bases its inference?

Our systematic analysis provides compelling evidence for the effectiveness of this approach. With improvements ranging from 10\% to 33\% across performance metrics and the ability to achieve near-reference performance for multiple measures, SFI-aware calibration represents a practical and powerful tool for clinical AI deployment. 

Perhaps most importantly, our findings demonstrate that diagnostic signal decay—long considered an intractable challenge in clinical AI—may be more amenable to systematic intervention than previously believed. The substantial performance recovery achieved through calibration alone suggests that current deployment strategies may be unnecessarily accepting of performance degradation.

In an era of widespread model deployment, portability cannot come at the cost of Precision. Diagnostic signal decay is a solvable problem if we are willing to measure it. In this context, SFI-aware calibration opens a door for explorations toward self-calibrating, fidelity-aware learning systems that maintain performance integrity across diverse clinical environments.

Future work should focus on real-world validation across multiple healthcare systems, exploration of SFI-aware training strategies, and developing automated calibration frameworks that can adapt to local data characteristics without manual parameter tuning. The foundation established here provides a roadmap for more robust and reliable clinical AI systems.

\section*{Declaration of Interests}
The authors declare no competing interests.

\section*{Acknowledgments}
This study has been supported by grants from the National Institute on Aging (NIA) RF1AG074372.

\section*{Data and Code Availability}
The R package associated with this study can be accessed at the following GitHub repository: https://github.com/clai-group/SFI

\pagebreak

\pagebreak

\appendix

\section*{Appendix A: Signal Fidelity Index Formulations} 

For patient $i$, the SFI is defined as:  
\[
\text{SFI}_i = \frac{1}{6} \left(
\text{Specificity}_i +
\text{TemporalConsistency}_i +
\text{Entropy}_i +
\text{ContextualConcordance}_i +
\text{MedicationAlignment}_i +
\text{TrajectoryStability}_i
\right).
\]

The six components are specified as follows:  

\[
\text{Specificity}_i = \frac{\text{Count of specific dementia codes}}{\text{Total dementia codes}}
\]

\[
\text{TemporalConsistency}_i = 1 - \frac{\text{Number of diagnosis changes}}{\text{Total encounters} - 1}
\]

\[
\text{Entropy}_i = 1 - \frac{-\sum_{j} p_j \log_2(p_j)}{\log_2 K}, \quad 
p_j = \text{proportion of diagnosis code } j,\; K = \text{number of unique codes}
\]

\[
\text{ContextualConcordance}_i = \frac{\text{Contextually appropriate diagnosis codes}}{\text{Total encounters}}
\]

\[
\text{MedicationAlignment}_i = \frac{\text{Encounters with both dementia code and dementia-specific medication}}{\text{Total encounters with dementia code}}
\]

\[
\text{TrajectoryStability}_i =
\begin{cases}
1, & \text{if most common inpatient code} = \text{most common outpatient code}, \\
0, & \text{otherwise}.
\end{cases}
\]

\section*{Appendix B: Mathematical Proof of SFI-Aware Calibration}

\subsection*{1. Setup and Definitions}

Let $\hat{y}_{i,\text{raw}} = P(Y=1 \mid X_i, D_S)$ be the predicted probability for patient $i$ based on a model trained on source dataset $D_S$. Define the Signal Fidelity Index (SFI) for each patient as $SFI_i$, and let $\bar{SFI}_S$ denote the average SFI in the training data.

\subsection*{2. Miscalibration and First-Order Approximation}

We hypothesize a function $f$ that maps raw predictions and fidelity to calibrated outputs:

\begin{align*}
\hat{y}_{i,\text{ideal}} = f(\hat{y}_{i,\text{raw}}, SFI_i)
\end{align*}

Approximating $f$ via first-order Taylor expansion around $\bar{SFI}_S$:

\begin{align*}
f(\hat{y}_{i,\text{raw}}, SFI_i) \approx f(\hat{y}_{i,\text{raw}}, \bar{SFI}_S) + (SFI_i - \bar{SFI}_S) \cdot \frac{\partial f}{\partial SFI}\bigg|_{SFI = \bar{SFI}_S}
\end{align*}

\subsection*{3. SFI-Aware Calibration Formula}

We model the adjustment as:

\begin{align*}
\hat{y}_{i,\text{calibrated}} = \hat{y}_{i,\text{raw}} \cdot \left[1 + \alpha \cdot \frac{SFI_i - \bar{SFI}_S}{\bar{SFI}_S} \right]
\end{align*}

where $\alpha$ is a tunable sensitivity parameter.

\subsection*{4. Calibration Error Improvement}

Let the calibration error be defined as the squared difference from the true label:

\begin{align*}
\Delta_{error} = E\left[(Y_i - \hat{y}_{i,\text{raw}})^2 - (Y_i - \hat{y}_{i,\text{calibrated}})^2\right]
\end{align*}

Expanding this:

\begin{align*}
\Delta_{error} = E\left[2(Y_i - \hat{y}_{i,\text{raw}})(\hat{y}_{i,\text{calibrated}} - \hat{y}_{i,\text{raw}}) - (\hat{y}_{i,\text{calibrated}} - \hat{y}_{i,\text{raw}})^2\right]
\end{align*}

Substitute the calibration formula:

\begin{align*}
\hat{y}_{i,\text{calibrated}} - \hat{y}_{i,\text{raw}} = \hat{y}_{i,\text{raw}} \cdot \alpha \cdot \frac{SFI_i - \bar{SFI}_S}{\bar{SFI}_S}
\end{align*}

\subsection*{5. Interpretation}

If $(Y_i - \hat{y}_{i,\text{raw}})$ and $(SFI_i - \bar{SFI}_S)$ are positively correlated, then the expected error difference $\Delta_{error}$ is positive, thus improving calibration.

\subsection*{6. Optimal Alpha}

To find optimal $\alpha$, minimize calibration error:

\begin{align*}
\alpha^* = \frac{E[(Y_i - \hat{y}_{i,\text{raw}}) \cdot \hat{y}_{i,\text{raw}} \cdot \frac{SFI_i - \bar{SFI}_S}{\bar{SFI}_S}]}{E[\hat{y}_{i,\text{raw}}^2 \cdot (\frac{SFI_i - \bar{SFI}_S}{\bar{SFI}_S})^2]}
\end{align*}

\appendix
\section*{Appendix C: Calibration Algorithm Details} 

\subsection*{Phase 1: Optimal Alpha Selection}

For each metric $m$ and alpha value $\alpha_i$, the mean improvement across testing datasets was calculated as:
\[
\Delta_m(\alpha_i) = \overline{P_{m,cal}(\alpha_i)} - \overline{P_{m,raw}}
\]
where $P_{m,cal}(\alpha_i)$ denotes calibrated performance at $\alpha_i$, and $P_{m,raw}$ denotes the raw (uncalibrated) performance.  

The optimal alpha for each metric was defined as:
\[
\alpha_{opt,m} = \min\{\alpha_i : p(\Delta_m(\alpha_i)) < 0.05 \;\; \text{and} \;\; p(\Delta_m(\alpha_{i+1})) \geq 0.05\},
\]
subject to the constraint $\alpha_{opt,m} \leq 2.0$ to prevent over-calibration.  

The overall recommended calibration parameter was calculated as:
\[
\alpha_{recommended} = \min\left(\text{median}\{\alpha_{opt,m}\}, 2.0\right).
\]

\subsection*{Phase 2: Batch-Level Performance Analysis} 

For each batch $b$ and metric $m$, raw and calibrated performance were aggregated as:
\[
\bar{P}_{m,raw}^{(b)} = \frac{1}{50}\sum_{j=1}^{50} P_{m,raw}^{(b,j)}, \quad
\bar{P}_{m,cal}^{(b)} = \frac{1}{50}\sum_{j=1}^{50} P_{m,cal}^{(b,j)},
\]
where $P_{m,raw}^{(b,j)}$ and $P_{m,cal}^{(b,j)}$ represent the raw and calibrated performance, respectively, for metric $m$ in batch $b$, dataset $j$.

\subsection*{Distance-to-Reference Analysis} 
To evaluate how calibration improved proximity to reference standards, we defined distance metrics as:

\[
d_{raw} = |\bar{P}_{m,raw} - \bar{P}_{m,ref}|, \quad
d_{cal} = |\bar{P}_{m,cal} - \bar{P}_{m,ref}|
\]

The improvement in proximity was quantified as:

\[
\text{Distance Improvement} = d_{raw} - d_{cal},
\]

and the percentage improvement relative to the raw model was given by:

\[
\text{Percent Closer to Reference} = \frac{d_{raw} - d_{cal}}{d_{raw}} \times 100\%.
\]

\end{document}